\documentclass[journal,onecolumn]{IEEEtran}
\usepackage{epsfig}
\usepackage{amsmath}
\usepackage{amssymb}
\usepackage[nolist]{acronym}
\usepackage{subcaption}
\usepackage{multirow}

\newacro{lstm}[LSTM]{Long Short-Term Memory}
\newacro{lsta}[LSTA]{Long Short-Term Attention}
\newacro{clstm}[ConvLSTM]{Convolutional Long Short-Term Memory}
\newacro{rnn}[RNN]{Recurrent Neural Network}
\newacro{cnn}[CNN]{Convolutional Neural Network}
\newacro{tsn}[TSN]{Temporal Segment Network}
\newacro{trn}[TRN]{Temporal Relation Network}
\newacro{tdn}[TDN]{Temporal Difference Network}
\newacro{mfn}[MFNet]{Motion Feature Network}
\newacro{fc}[FC]{Fully Connected}
\newacro{bn}[BN]{Batch Normalization}
\newacro{gru}[GRU]{Gated Recurrent Unit}
\newacro{sgd}[SGD]{Stochastic Gradient Descent}

\title{Hierarchical Feature Aggregation Networks for Video Action Recognition}

\author{Swathikiran Sudhakaran$^{1,2}$, Sergio Escalera$^{3,4}$, Oswald Lanz$^{1}$\\ 
	$^{1}$Fondazione Bruno Kessler, Trento, Italy\\
	$^{2}$University of Trento, Trento, Italy\\
	$^{3}$Computer Vision Center, Barcelona, Spain\\
	$^{4}$University of Barcelona, Barcelona, Spain\\
	{\tt\small \{sudhakaran,lanz\}@fbk.eu, \tt\small sergio@maia.ub.es}
}

\def\etal{\emph{et al}.}

\begin{document}

\maketitle

\begin{abstract}
	Most action recognition methods base on a) a late aggregation of frame level CNN features using average pooling, max pooling, or RNN, among others, or b) spatio-temporal aggregation via 3D convolutions. The first assume independence among frame features up to a certain level of abstraction and then perform higher-level aggregation, while the second extracts spatio-temporal features from grouped frames as early fusion. In this paper we explore the space in between these two, by letting adjacent feature branches interact as they develop into the higher level representation. The interaction happens between feature differencing and averaging at each level of the hierarchy, and it has convolutional structure that learns to select the appropriate mode locally in contrast to previous works that impose one of the modes globally (e.g. feature differencing) as a design choice. We further constrain this interaction to be conservative, e.g. a local feature subtraction in one branch is compensated by the addition on another, such that the total feature flow is preserved. We evaluate the performance of our proposal on a number of existing models, i.e. TSN, TRN and ECO, to show its flexibility and effectiveness in improving action recognition performance.

	
\end{abstract}

\section{Introduction}
\label{sec:intro}
Video action recognition requires strong temporal reasoning.
Temporal reasoning in deep networks can be implemented by 3D (space+time) convolutions, temporal (average) pooling, or recurrent layers that aggregate frame-level or spatio-temporal representations of an input sequence. The effectiveness of temporal reasoning hereby depends on the representation upon which the aggregation is performed. 

While spatio-temporal features are more effective than frame-level features in video recognition tasks, they are also more costly to compute and harder to train, including a large number of parameters. Furthermore, even with a deep backbone the receptive field of spatio-temporal features are confined to the small fixed-size frame window from which they were computed. Since more abstract representations benefit from larger temporal context, a gradual increase of the temporal extend of receptive fields in the higher layers of the feature extraction backbone would be desired instead.

A feasible solution to increase the temporal size of receptive fields is to let features from adjacent branches interact. \ac{tsn}~\cite{tsn} showed that it is useful to apply the difference of frames as input to a CNN for video action recognition. \ac{tdn}~\cite{tdn} extended this idea to compute the difference of frame level features obtained from each layer in a CNN. Such hardcoded approach considers that there is a significant variation of information in adjacent frames. This is however not always guaranteed in practice. It is therefore desirable to let a network decide when to subtract the features from adjacent frames.


\begin{figure}
    \centering
    \begin{subfigure}[b]{0.35\textwidth}
		\includegraphics[scale=0.3]{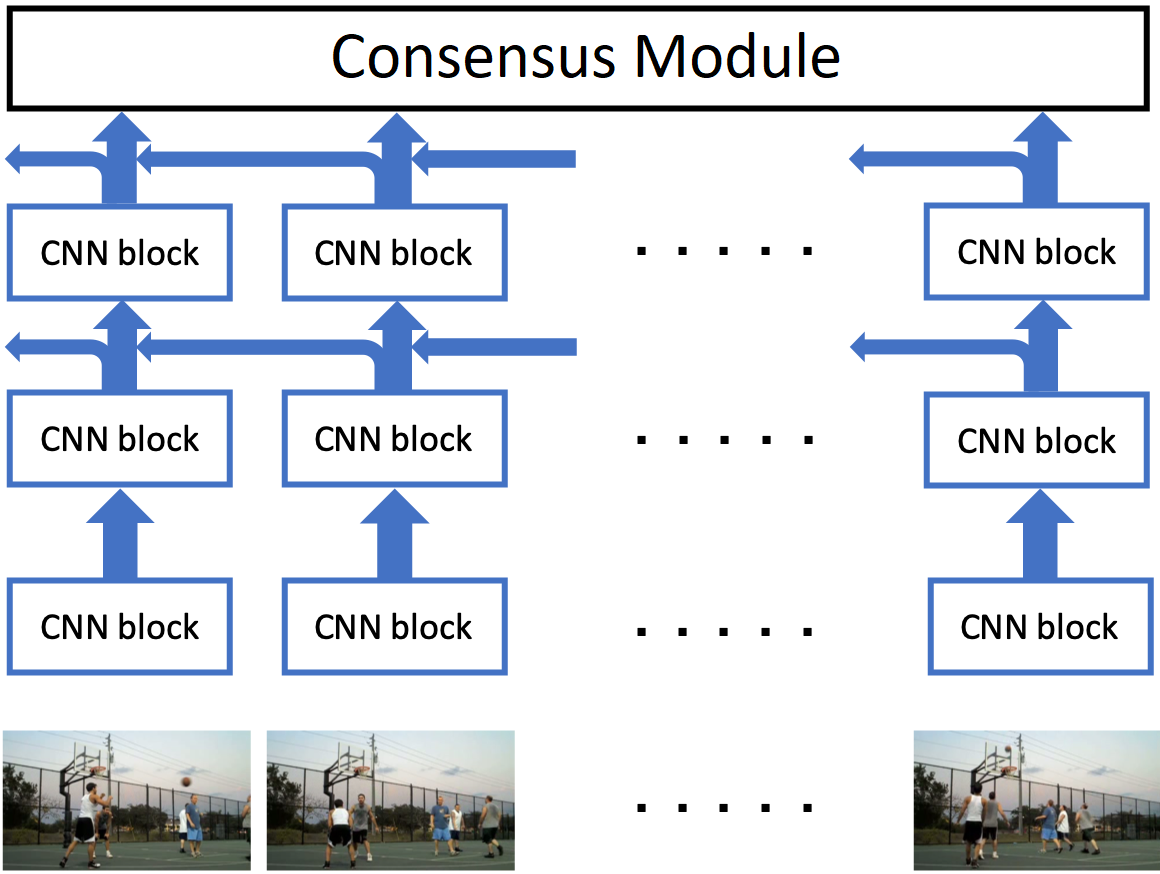}
		\caption{HF-Net}
		\label{fig:hfnet} 
	\end{subfigure} \hspace{3cm}
	\begin{subfigure}[b]{0.3\textwidth}
		\includegraphics[scale=0.42]{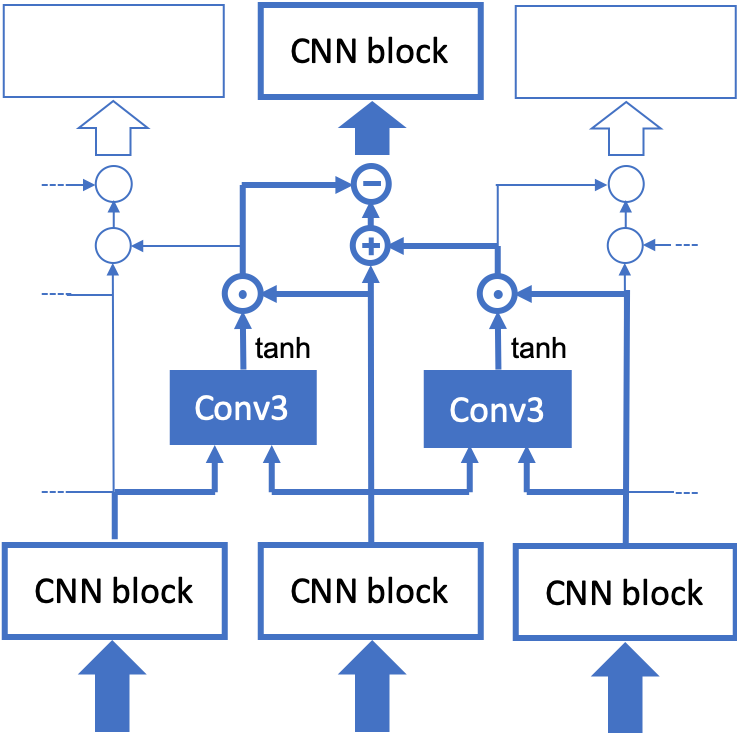}
		\caption{HF module}
		\label{fig:hfmod}
	\end{subfigure}
	\caption{HF-Net: We introduce hierarchical feature aggregation in video deep architectures with CNN backbone. At each layer the feature from a CNN block is gated and its residual is transferred to the adjacent branch. Gating is learnt jointly with the network during training, and decides locally whether, and to what extend, a feature differencing or feature averaging is performed. The temporal extend of receptive fields increases with network depth, which helps developing spatio-temporal representations without 
	expensive 3D convolutions.
}
	\label{fig:basic_block}
\end{figure}

In this paper we present a hierarchical feature aggregation scheme that is lightweight and can be plugged into any deep architecture with CNN backbone. In particular, when plugged into \ac{tsn}, a performance gain of 24.2\% is obtained on Something-v1~\cite{goyal2017something} dataset with an addition of only 1.14\% parameters and 1.04\% Floating Point Operations (FLOPs). The main idea, as illustrated in Fig.~\ref{fig:hfnet}, is to transfer features between adjacent branches at each layer in a way that adjacent features interact as they develop into the higher level representation. The amount of feature transfer is controlled through a convolutional gate that decides to what amount a feature is subtracted or added into the adjacent branch, as detailed in Fig.~\ref{fig:hfmod}. Through end-to-end training, the network learns to route features through the hierarchy, hereby causing the temporal extend of receptive fields to grow with depth and adapt to the input. We evaluate our scheme on a number of existing models, TSN, TRN and ECO, and show its flexibility and effectiveness in improving action recognition performance.

The rest of the paper is organized as follows. Sec.~\ref{sec:related_works} reviews work on action recognition most related to our proposal. Sec.~\ref{sec:hfnet_main} presents the hierarchical feature aggregation network. Experimental results are reported in Sec.~\ref{sec:results} and Sec.~\ref{sec:conclusions} concludes the paper.





\section{Related work}

\label{sec:related_works}

Inspired by the performance improvements obtained by CNNs on image recognition~\cite{resnet, szegedy2016rethinking}, many deep learning based techniques have been developed for video action recognition. The most effective and simple extension was developed by Simonyan and Zisserman~\cite{twoStream}. Their method consists of two different CNNs trained on a single RGB image frame and a stack of optical flow images followed by a late fusion of the prediction scores. The image stream encodes the appearance information while the optical flow stream encodes the motion information. Several works followed this approach to find a suitable fusion of the two streams~\cite{feichtenhofer2016convolutional} and exploring residual connections between them~\cite{feichtenhofer2016spatiotemporal}. The downside of this approach is its reliance on externally computed optical flow which is computationally intensive.

In order to address the aforementioned problem, researchers have explored techniques for extracting spatio-temporal features from the RGB frames itself. Karpathy~\etal~\cite{karpathy2014large} examined several fusion approaches at various levels of the CNN hierarchy. They found that a slow fusion approach where the features from adjacent frames are fused at multiple hierarchical levels of the CNN results in the best performance. Their fusion approach stacks the convolutional features from adjacent frames and perform temporal convolutions for extracting the temporal features. Later, Tran~\etal~\cite{tran2015learning} developed 3DCNN with 3D convolutional layers so that spatio-temporal features can be extracted from a set of multiple RGB frames. Their approach showed that 3D convolutions are capable of extracting spatio-temporal features from small video segments consisting of a small number of frames. Several approaches have been later proposed for exploiting the parameters learned by a 2DCNN for image recognition~\cite{carreira2017quo, qiu2017learning, tran2018closer}. Carreira and Zisserman~\cite{carreira2017quo} developed a 3DCNN by inflating the 2D filters to 3D. Qiu~\etal~\cite{qiu2017learning} and Tran~\etal~\cite{tran2018closer} proposed to factorize the 3D convolutions to a 2D convolution for spatial encoding followed by a 1D convolution for temporal encoding. Wang~\etal developed an architecture based on 3DCNNs which decouples the spatial and temporal encoding. Their approach performs a linear encoding on individual frame features and a multiplicative encoding between a set of features from multiple frames. Such approaches using 3DCNNs have two major drawbacks, firstly, the massive increase in the number of trainable parameters and secondly, they extract spatio-temporal features from just a small sample of adjacent frames. The first problem makes such approaches difficult to be trained on smaller datasets and the second problem makes them incapable of extracting long range spatio-temporal features from videos.

In order to extract long-range spatio-temporal features, several techniques that perform sparse sampling of the video frames, as opposed to the dense sampling used in 3DCNNs, have been proposed. Such approaches use a 2DCNN for extracting the frame level features followed by a late fusion using \acp{rnn}~\cite{yue2015beyond, lrcn}, pooling techniques such as average pooling or max pooling~\cite{tsn}, \ac{fc} layers~\cite{trn}, or 3D convolutions~\cite{eco}, among others. Such late fusion based approaches ignore the information extracted at the different layers of a CNN. Several approaches have been proposed to address this drawback by fusing the features from consecutive frames at different layers of hierarchy of a CNN~\cite{tdn, off, lee2018motion}. Ng and Davis~\cite{tdn} propose to use an extra CNN that accepts the difference of feature maps from adjacent frames at different layers of a separate CNN. The final prediction is done by average pooling of the scores obtained from the two networks. Sun~\etal improves this approach by applying a Sobel filter on the features in addition to the temporal differencing operation. Lee~\etal~\cite{lee2018motion} improves this method by forwarding the spatial and temporal features across a single network. This is achieved by applying a set of fixed filters followed by differencing of the features from adjacent frames.
Lin~\etal~\cite{tsm} proposes a plug and play module that shifts the channels in the feature tensor across the temporal dimension for transferring information across frames.

Our approach is related to~\cite{tdn}, \cite{off}, \cite{lee2018motion} and \cite{tsm}. These approaches perform a fixed interaction between feature maps from adjacent frames whereas our approach learns a set of filters that can perform point wise difference or average operations across the feature maps.

\section{HF-Net: Hierarchical Feature Aggregation Networks}
\label{sec:hfnet_main}
In this section we describe our hierarchical feature aggregation for video representation learning architectures that utilize a CNN backbone for frame-level or spatio-temporal feature extraction. We then present details of our action recognition model used in the experiments.

\subsection{Hierarchical Feature Aggregation}
\label{sec:hfmodule}

Deep architectures with CNN backbone by design do not account for correlations that
may exist between frame-level or spatio-temporal features at the earlier layers of the CNN.
Indeed, in a CNN backbone a feature $F_{t,l}$ for input $t$ at layer $l$ is computed from $F_{t,l-1}$ independently of the features from inputs other than $t$, that is, $F_{t,l} = \verb+block+(F_{t,l-1})$, where \verb+block+ represents a set of convolutional layers with non-linearities. This may limit performance by design when the input sequence exhibits strong temporal dependence such as video frames in action recognition. In such a setting, a feature aggregation in the early layers may better capture the temporal dependencies in the data sequence.

Our hierarchical feature aggregation is based on the assumption that adjacent features will benefit from interacting in order to produce a more abstract representation at each layer of the architecture. We want this interaction to be pairwise and feed-forward computable, that is, our backbone hierarchical aggregation scheme is realized with a re-designed $F_{t,l} = \verb+block2+(F_{t,l-1},F_{t-1,l-1})$.

In order to define \verb+block2+ we consider two elements that are pervasive in the state-of-the-art architectures. First, most architectures perform feature pooling at the higher level, most often, average pooling. Second, many networks follow a two stream structure where optical flow is late fused from a separate branch. Some works~\cite{tsn, tdn} already reported performance improvements even with frame differencing as a simplified version of optical flow. In our design of \verb+block2+ both feature differencing and feature averaging are considered. We want the network to select or interpolate between these two modes, and maximise flexibility by learning to decide so locally in the feature tensor. This way, we provide the network with the capability to selectively route features through the hierarchy and build up discriminative representations integrating temporal context. That is, growing temporal receptive fields as we go up in the hierarchy. 

We define \verb+block2+ as:

\begin{equation}
F_{t,l} = F'_{t,l} + G_{t,l} \odot F'_{t+1, l} - G_{t-1,l} \odot F'_{t, l}, \label{eq:new_feat_t+1}
\end{equation}
where $F'_{t,l} = \verb+block+(F_{t,l-1})$, `$*$' and `$\odot$' represent convolution operation and Hadamard product, respectively, and $G_t$ acts as a gating tensor that chooses between averaging and differencing operations. Fig.~\ref{fig:hfmod} illustrates \verb+block2+. Note that we subtract a gated $G_{t-1,l} \odot F'_{t,l}$ from the backbone feature $F'_{t,l}$ but we add it when computing $F_{t-1,l}$. Thus, the total feature flow on the sequence is preserved. Gating $G_t$ is generated by performing a 3D convolution on the stacked features with a $2\times3\times3$ kernel followed by $\tanh$ non-linearity:

\begin{equation}
G_{t,l} = \tanh(W_l * [F'_{t,l} , F'_{t+1, l}] + B_{l}). \label{eq:gate}
\end{equation}
Since the range of $\tanh$ non-linearity is [-1, 1], the network is capable of selecting from both averaging and differencing operations. 
$W_l$ and $B_l$ represent the weights and the bias of the 3D convolution. We use a $2\times3\times3$ kernel with $1$ output channel in the 3D convolution. 






\subsection{Layer Implementation}

The proposed hierarchical aggregation across each layer can be realized in parallel, thereby allowing for a fast training and inference. Assuming that sequences of frames are applied as a batch to the network, the parallel implementation of hierarchical aggregation is realized as:

\begin{eqnarray}
F^{\verb:+:} &=& \verb+block+(F_{:,l-1}),\\
F^{\verb:-:} &=& \verb+shift_left+(F^{\verb:+:}), \\
G^{\verb:-:} &=& \tanh(\verb+conv3+([F^{\verb:+:},F^{\verb:-:}])), \\
G^{\verb:+:} &=& \verb+shift_right+(G^{\verb:-:}), \\
F_{:,l} &=& G^{\verb:-:} \odot F^{\verb:-:} + (1- G^{\verb:+:}) \odot F^{\verb:+:}.
\end{eqnarray}


At layer $l$ of the backbone, we first feed the input sequence $F_{:,l-1}$ through the corresponding convolutional \verb+block+ of the backbone to obtain a new sequence $F^{\verb:+:}$. We then left-shift and zero-pad to obtain $F^{\verb:-:}$ and compute the gating tensor $G^{\verb:-:}$ through a 3D convolution on the stacked $[F^{\verb:+:},F^{\verb:-:}]$. This is followed by $\tanh$ to map the resulting tensor to values in the range [-1, 1]. In order to obtain the sequence $F_{:,l}$ of output features, the Hadamard product of the gating tensor $G^{\verb:-:}$ and the left shifted feature tensor $F^{\verb:-:}$ is added to the input feature tensor $F^{\verb:+:}$ while the Hadamard product of the right shifted gating tensor $G^{\verb:+:}$ and the input feature tensor $F^{\verb:+:}$ is subtracted.

\subsection{Hierarchical Feature Aggregation Networks}
\label{sec:hfanet}
Hierarchical Feature Aggregation Network (HF-Net) is obtained by adding the Hierarchical Feature Aggregation Module, presented in Sec.~\ref{sec:hfmodule}, at various levels of the backbone CNN used in the network. 
Our approach is generalizable to any CNN architecture (e.g, HF-TSN). In the experiments, we use Inception with \ac{bn}~\cite{inception_bn} as our backbone CNN. We plug in our Hierarchical Feature Aggregation module after each Inception block in the CNN. At each iteration, a set of sparsely sampled frames $K$ from a video is passed to the network. In addition to the common feed forward flow of information in conventional CNNs, the hierarchical feature aggregation modules enable a horizontal flow of information across the features at different levels of the CNN hierarchy. The output features corresponding to all the frames from the final layer of the CNN can then be pooled together for encoding the long range temporal features. Our architecture is highly flexible and can be combined with a number of temporal pooling techniques such as \ac{tsn}~\cite{tsn}, \ac{trn}~\cite{trn}, \ac{gru}~\cite{dwibedi2018temporal}, or 3DCNN~\cite{eco}. In Sec.~\ref{sec:ablation}, we evaluate the performance of our proposal on a number of existing approaches and show its flexibility and effectiveness in improving action recognition performance. 


\section{Experiments and Results}
\label{sec:results}
We evaluate the proposed hierarchical feature aggregation strategy on a number of existing models and standard action recognition datasets. We first briefly describe the considered datasets, evaluation metrics, and implementation details. Then, we perform an ablation analysis to quantitatively evaluate the impact of different HF properties and components. Finally, we compare the performance of various HF-Net architectures against state-of-the-art results on different public action recognition datasets. We show HF enhances recognition performance of all models where it is applied, with negligible additional complexity.

\subsection{Datasets and evaluation protocol}

\begin{figure}[t]
	\centering
		\begin{subfigure}[b]{0.3\textwidth}
		\includegraphics[width=35px,height=25px]{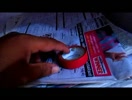}
		\includegraphics[width=35px,height=25px]{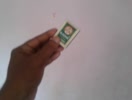}
		\includegraphics[width=35px,height=25px]{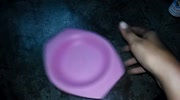}
		\caption{Something-v1}
		\label{fig:something}
		\end{subfigure}\hskip3mm
		\begin{subfigure}[b]{0.3\textwidth}
    	\includegraphics[width=35px,height=25px]{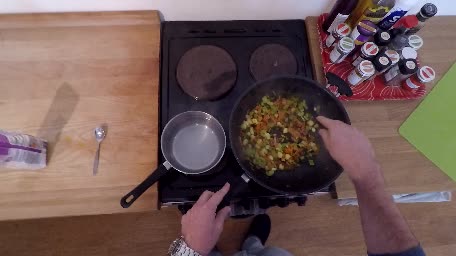}
    	\includegraphics[width=35px,height=25px]{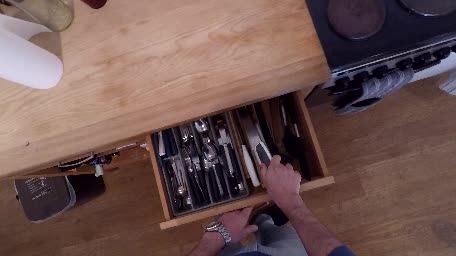}
    	\includegraphics[width=35px,height=25px]{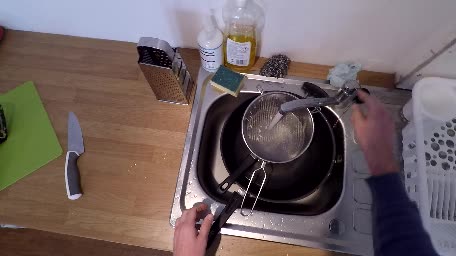}
    	\caption{EPIC-KITCHENS}
    	\label{fig:epic}
    	\end{subfigure}\hskip3mm
		\begin{subfigure}[b]{0.3\textwidth}
		\includegraphics[width=35px,height=25px]{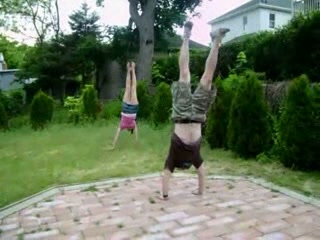}
		\includegraphics[width=35px,height=25px]{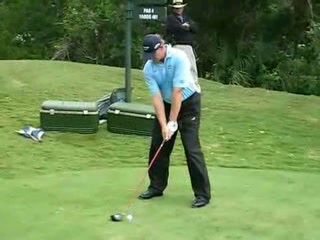}
		\includegraphics[width=35px,height=25px]{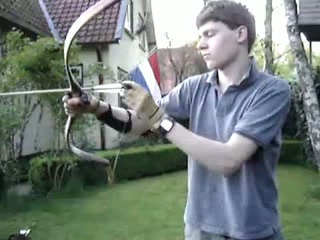}
		\caption{HMDB51}
		\label{fig:hmdb}
		\end{subfigure}
		\caption{Samples from the datasets used for evaluating the performance of HF-Nets.}
	\label{fig:dataset_samples}
\end{figure}

We evaluate our proposed hierarchical feature aggregation technique on three standard action recognition benchmarks, Something-v1~\cite{goyal2017something}, EPIC-KITCHENS~\cite{damen2018scaling} and HMDB51~\cite{kuehne2013hmdb51}.  Something-v1 consists of 86K and 11K videos in the training and validation sets from 174 action classes. We report the performance on the validation set. 
EPIC-KITCHENS dataset comprises of egocentric videos with fine-grained activity labels. The labels are provided as verb and nouns and the dataset consists of 24K videos in the training set and 10K videos in the test set. We report the performance obtained on the test set. 
HMDB51 dataset consists of videos collected from Youtube and contains around 6000 video clips from 51 action categories. The dataset is provided with three standard train/test splits and the final recognition accuracy is reported as the average of the accuracy obtained on the three splits.
Both Something-v1 and EPIC-KITCHENS datasets consists of crowd collected videos with actions involving objects. Something-v1 gives importance to the actions rather than the objects involved in the action while EPIC-KITCHENS gives relevance to both actions and objects. HMDB51 is a smaller dataset with less complex action categories of shorter temporal span, which can be identified with simple appearance cues. Some sample frames from the datasets are shown in Fig.~\ref{fig:dataset_samples}. In addition to the recognition accuracy, we also compare the complexity of the models in terms of number of parameters and Floating Point Operations (FLOPs). 



\subsection{Implementation Details}

As explained in Sec.~\ref{sec:hfanet}, we choose BNInception as the backbone. The proposed HF module is added after each Inception block of the CNN. The entire network, including the \ac{bn} layers, is trained for 60 epochs using \ac{sgd} optimization algorithm with a batch size of 32. The learning rate is fixed as 0.001 and is reduced by a factor of 0.1 after 25 and 40 epochs. Dropout at a rate of 0.5 is applied before the final classification layer to avoid overfitting. Random scaling and cropping, as proposed in~\cite{tsn} is used as data augmentation. During inference, only the center crop of the frames are used. In all experiments, we use 16 frames as the input to the model. 


\begin{table}[t]
	\centering 
	\begin{tabular}{|l|c|c|c|}
		\hline
		\textbf{Model} & \textbf{Accuracy (\%)} & \textbf{Params} & \textbf{FLOPs}\\   \hline
		Baseline & 17.77 & 10.47M & 32.73G\\ \hline
		Baseline+1HF & 19.37 & 10.47M & 32.81G \\ \hline
		Baseline+5HF & 34.15 & 10.53M & 32.90G \\ \hline
		Baseline+10HF & 41.97 & 10.59M & 33.07G \\ \hline
		Baseline+10HF (non-Conservative) & 38.81 & 10.69M & 33.47G\\ \hline
	\end{tabular}
	\caption{Ablation analysis conducted on the validation set of Something-v1 dataset.
    }
	\label{tab:ablation}
\end{table}

\begin{figure}[t]
	\centering
		\begin{subfigure}[b]{1\textwidth}
			\hspace{2cm}
		\includegraphics[angle=90, scale=0.35]{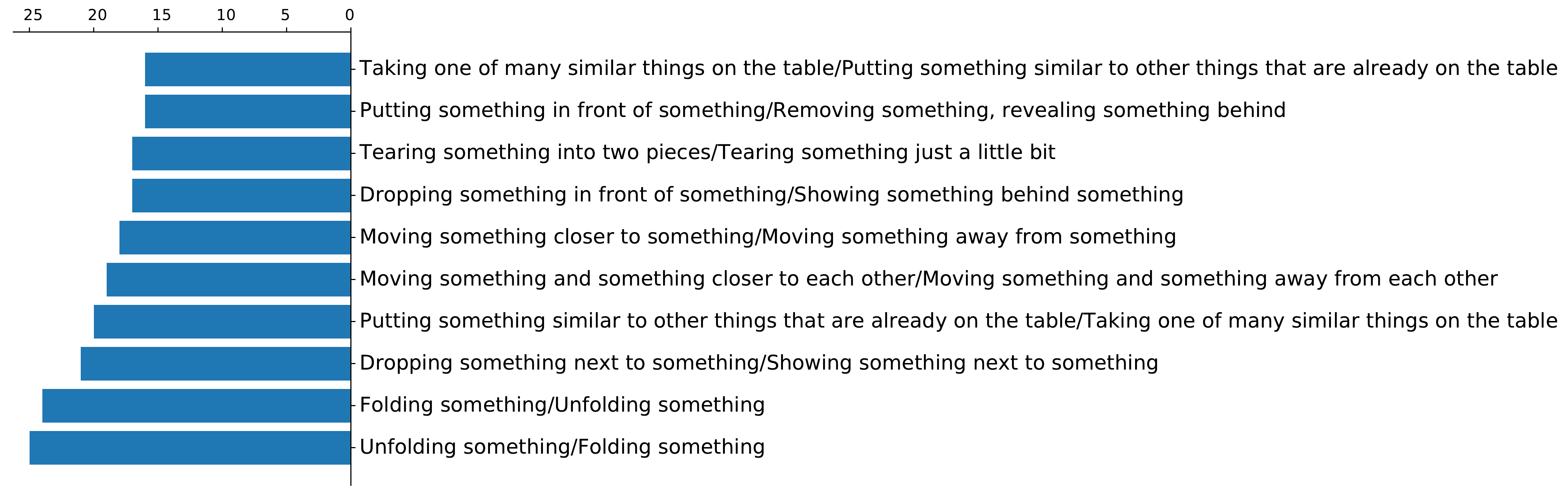}
		\caption{Improved classes over \ac{tsn}}
		\label{fig:classes_imp}
	\end{subfigure} \\ \vspace*{.5cm}
	\begin{subfigure}[b]{0.35\textwidth}
		\includegraphics[scale=0.4]{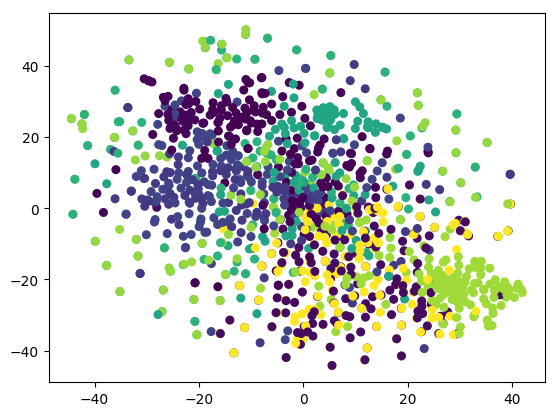}
		\caption{\ac{tsn} t-SNE plot}
		\label{fig:tsn_tsne}
	\end{subfigure} \hspace{3cm}
	\begin{subfigure}[b]{0.35\textwidth}
		\includegraphics[scale=0.4]{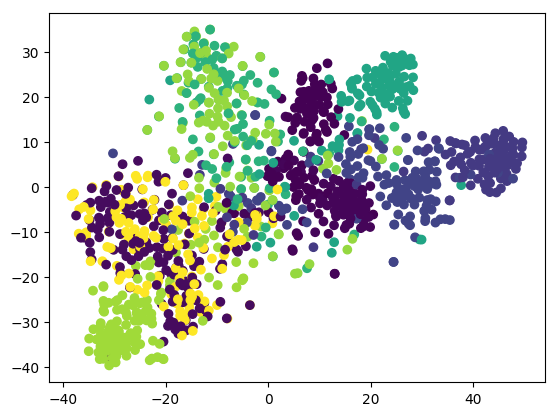}
		\caption{HF-\ac{tsn} t-SNE plot}
		\label{fig:hftsn_tsne}
	\end{subfigure}
		\caption{(a) Action classes in Something-v1 dataset with the highest improvement over \ac{tsn} by adding HF on the backbone. Y-axis labels are in the format true label (HF-TSN)/predicted label (TSN). X-axis shows the number of corrected samples for each class. (b) and (c) show the t-SNE plots of the output layer features preceding the \ac{fc} layer. The 10 improved classes shown in (a) are visualized in the t-SNE plots.}
	\label{fig:tsne_plots}

\end{figure}

\subsection{Ablation Analysis}
\label{sec:ablation}

In this section, we report the ablation analysis performed on the validation set of Something-v1 dataset. We compare the performance improvement by adding the proposed HF module on the CNN backbone of a standard action recognition technique. We choose \ac{tsn}~\cite{tsn} as the baseline since it is one of the standard techniques for video action recognition. \ac{tsn} divides each video into a pre-defined number of segments and applies one frame from each of the segments as the input to the network. The average of the output from each of the frames is used for computing the prediction scores. Thus, \ac{tsn} fails to encode the temporal relations between video frames and hence acts as a suitable baseline for showing the capability of the proposed hierarchical feature aggregation approach in extracting spatio-temporal features. 


Tab.~\ref{tab:ablation} shows the result of the ablation study conducted. We first evaluated the performance of the model when a single module is added after the final inception block of the backbone. We obtained an improvement of $1.6\%$. A further gain of $14.78\%$ is obtained by adding 5 modules after each of the final 5 inception blocks. Finally, we add 10 modules after each of the inception blocks which resulted in an improvement of $24.2\%$ over the \ac{tsn} baseline. We also evaluated the performance of the model when two independent 3D convolutional layers are used to compute the gating, thereby breaking the conservative flow of features. This increases the number of parameters and complexity slightly, albeit the performance of the network is reduced thereby proving that feature flow conservation is useful for spatio-temporal feature extraction in HF-Nets.


In Fig.~\ref{fig:classes_imp}, we show the top 10 action classes that improved the most by adding HF module to the backbone CNN of \ac{tsn}. From the figure, it can be seen that the network has enhanced its ability to distinguish between action classes that are similar in appearance, such as \verb+Unfolding something+ and \verb+Folding something+, \verb+Dropping something next+ \verb+to something+ and \verb+Showing something next+ \verb+to something+, etc. 
The t-SNE plots of the features from the final layer of the CNN corresponding to these 10 action classes are shown in Fig.~\ref{fig:tsn_tsne} and~\ref{fig:hftsn_tsne}. It can be seen that the features from HF-\ac{tsn} show a lower intra-class and higher inter-class variability compared to those from \ac{tsn}.

\begin{table}[h]
	\centering 
	\begin{tabular}{|l|c|c|c|c|}
		\hline
	\multirow{2}{*}{\textbf{Model}} & \multirow{2}{*}{\textbf{Backbone}} & \multirow{2}{*}{\textbf{Pre-training}} & \multicolumn{2}{c|}{\textbf{Accuracy (\%)}} \\ \cline{4-5}
	& & & Something-v1 & HMDB51\\ \hline
	    TDN~\cite{tdn} & ResNet-50 & ImageNet & - & 55.5 \\ \hline
		ECO~\cite{eco} & BNInception+3D ResNet-18 & ImageNet+Kinetics & 41.4 & 68.2\\ \hline
		MFNet-C~\cite{lee2018motion} & ResNet-101 & Scratch & 43.92 & -\\ \hline
		ARTNet~\cite{wang2018appearance} & 3D ResNet-18 & Kinetics & - & 70.9 \\ \hline \hline
		I3D~\cite{carreira2017quo} & 3D ResNet-50 & ImageNet+Kinetics & 41.6 & 74.8\\ \hline
		C3D~\cite{tran2015learning} & 3D ResNet-18 & Sports-1M & - & 62.1 \\ \hline
		3D-ResNeXt-101~\cite{hara2018can} & ResNeXt-101 & Kinetics & - & 70.2 \\ \hline
		R(2+1)D~\cite{tran2018closer} & 3D ResNet-34 & ImageNet+Kinetics & - & 74.5 \\ \hline
		Non-local I3D~\cite{wang2018videos} & 3D ResNet-50 & ImageNet+Kinetics & 44.4 & -\\ \hline
		Non-local I3D+GCN~\cite{wang2018videos} & 3D ResNet-50+GCN & ImageNet+Kinetics & 46.1 & -\\ \hline
		TSM~\cite{tsm} & ResNet-50 & ImageNet+Kinetics & 44.8 & 73.2\\ \hline
		S3D-G~\cite{xie2018rethinking} & Inception-V1 & ImageNet & 48.2 & -\\ \hline \hline
		TSN~\cite{tsn} & BNInception & ImageNet & 17.77 & 53.7 \\ \hline
		TRN~\cite{trn} & BNInception & ImageNet & 31.32 & 54.93 \\ \hline 
		ECOLite~\cite{eco} & BNInception+3D ResNet-18 & ImageNet+Kinetics & 42.2 & 68.5\\ \hline \hline
		HF-TSN & BNInception & ImageNet & 41.97 & 55.92 \\ \hline
		HF-TRN & BNInception & ImageNet & 39.17 & 57.61 \\ \hline
		HF-ECOLite & BNInception+3D ResNet-18 & ImageNet+Kinetics & - & 71.13\\ \hline
	\end{tabular}
	\caption{Comparison of proposed method with state-of-the-art techniques on the validation set of something-v1 and HMDB51 datasets.}
	\label{tab:something_sota}
\end{table}

\begin{figure}[t]
    \centering
    \begin{subfigure}[b]{.45\textwidth}\hspace{-6.5mm}
    \includegraphics[scale=0.08]{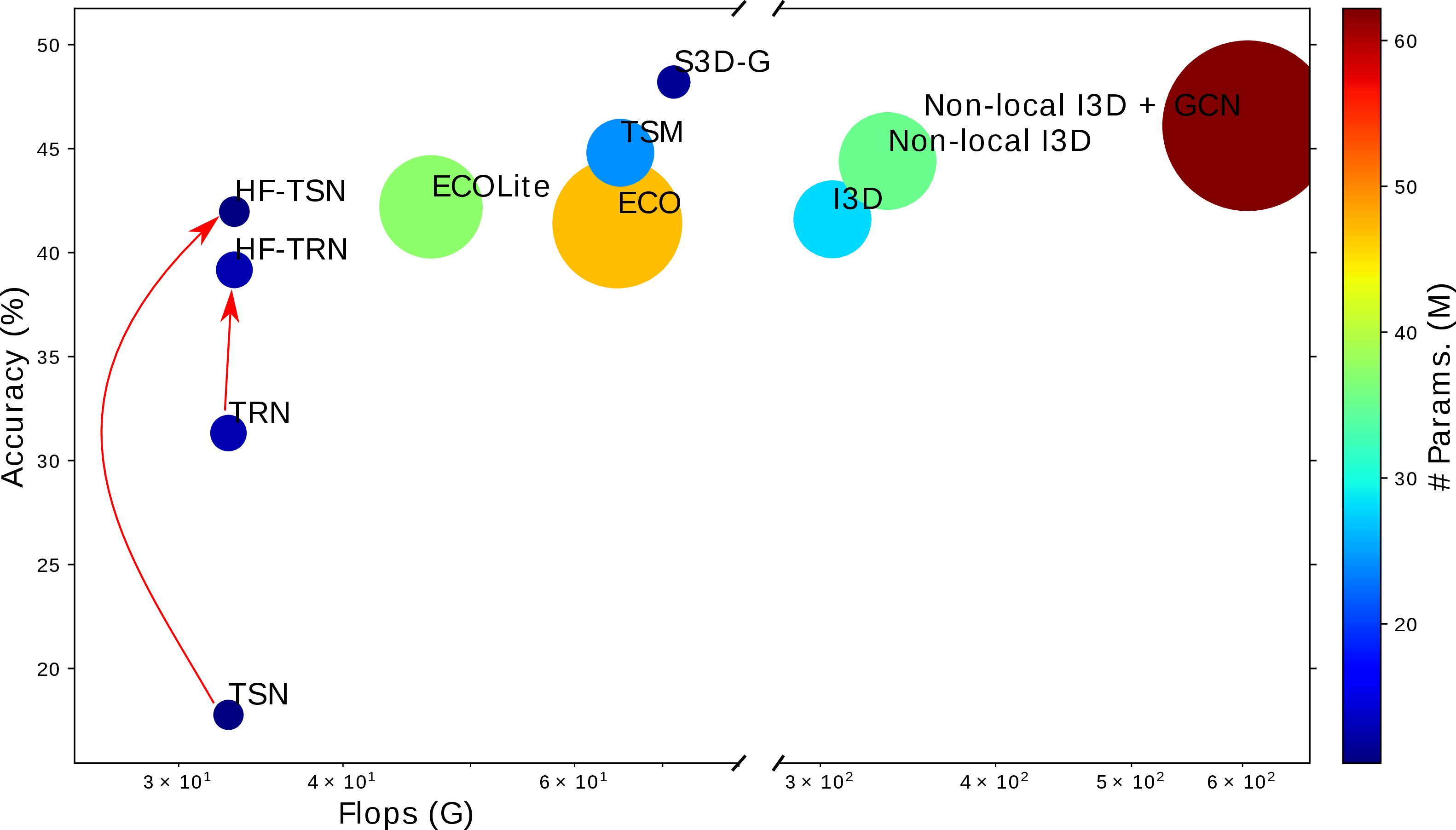}
    \caption{Something-v1 dataset}
    \label{fig:complexity_something}
    \end{subfigure}
    \begin{subfigure}[b]{.45\textwidth}\hspace{1.5mm}
    \includegraphics[scale=0.08]{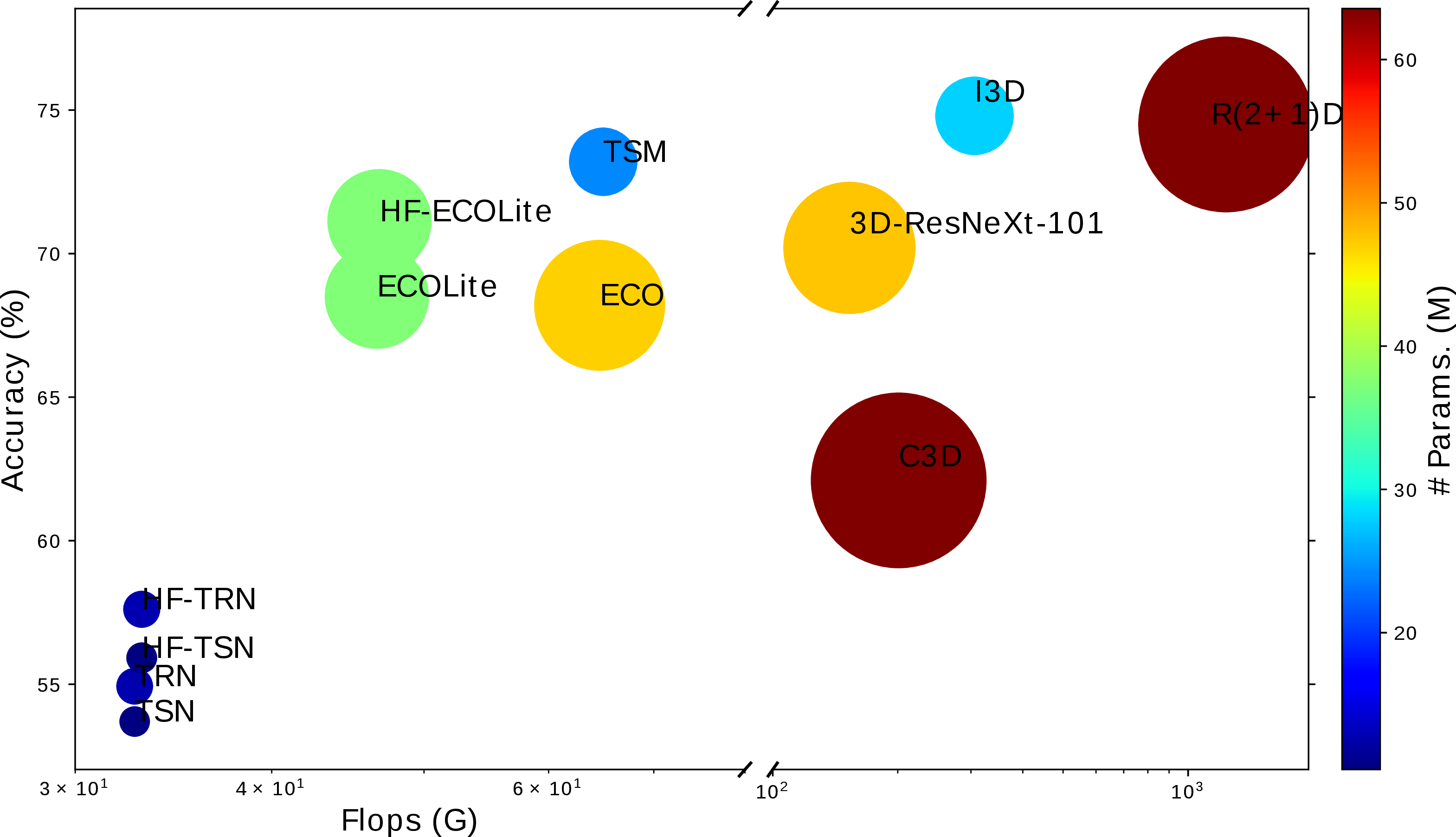}
    \caption{HMDB51 dataset}
    \label{fig:complexity_hmdb}
    \end{subfigure}
    \caption{Recognition accuracy (\%) of various state-of-the-art techniques on (a) Something-v1 and (b) HMDB51, plotted against the computational complexity in terms of GFLOPs. Color of the data point indicates the number of parameters (M, in millions).}
    \label{fig:complexity}
\end{figure}

\subsection{State-of-the-Art Comparison}

In order to compare methods at the same conditions, we compare only with those models using raw RGB frames as input. However, note that the proposed approach is extendable to optical flow images as well.

{\noindent \textbf{Something-v1 and HMDB51:}} Tab.\ref{tab:something_sota} compares the proposed approach with state-of-the-art techniques on Something-v1 and HMDB51 datasets. For Something-v1 dataset, in addition to \ac{tsn}, we also evaluate the performance of the proposed approach on \ac{trn}. In both methods, by adding the proposed HF module to the backbone, a large improvement in the performance is observed. From the table, one can see that the proposed approach results in comparable performance to other approaches that use a bigger backbone CNN such as ResNet-50 or I3D. It should also be noted that other methods which achieve superior performance use strong pre-training using Kinetics dataset. Importantly, the HF augmented models achieve this improved performance at a fraction of the number of parameters and FLOPs compared to other methods, allowing for faster inference and smaller memory footprint, e.g. rendering them to be deployed in mobile devices. Fig.~\ref{fig:complexity_something} illustrates the accuracy vs complexity of state-of-the-art techniques. From the figure, it can be seen that HF-Net results in a large boost in the recognition accuracy over the baseline models. 

For HMDB51, we augment three baselines, \ac{tsn}~\cite{tsn}, \ac{trn}~\cite{trn} and ECOLite~\cite{eco}, with HF module and observe a gain of more than $2\%$ on recognition accuracy over all the three baselines. As mentioned previously, HMDB51 consists of actions with shorter temporal duration. As a result, 3D CNNs that perform dense sampling of frames resulted in the best performance on this dataset~\cite{tran2018closer, carreira2017quo} due to the ability of 3D convolution layers in extracting spatio-temporal features from a short temporal window. ECOLite consists of a smaller number of 3D convolution layers on top of a 2D CNN backbone and thus is a middle ground between \ac{tsn} and 3D CNNs.
Even though ECOLite uses a more powerful (3D convolution) consensus layer than TSN (average pooling) and TRN (fully connected) the spatio-temporal features provided by HF aggregation are found to be beneficial. 
From the plot showing accuracy against complexity comparison of state-of-the-art approaches shown in Fig.~\ref{fig:complexity_hmdb}, one can see that HF-ECOLite performs on par with the bigger 3D CNN models. 

\begin{table}[t]
	\centering 
	\begin{tabular}{|c|l|c c c|c c c| c c c|c c c|}
		\hline
		& \textbf{Method} & \multicolumn{3}{c|}{\textbf{Top-1 Accuracy (\%)}} & \multicolumn{3}{c|}{\textbf{Top-5 Accuracy (\%)}} & \multicolumn{3}{c|}{\textbf{Precision (\%)}} & \multicolumn{3}{c|}{\textbf{Recall (\%)}} \\
		\cline{3-14}
		& & \parbox{0.7cm}{Verb} & \parbox{0.7cm}{Noun} & \parbox{0.8cm}{Action} & \parbox{0.7cm}{Verb} & \parbox{0.7cm}{Noun} & \parbox{0.8cm}{Action} & \parbox{0.7cm}{Verb} & \parbox{0.7cm}{Noun} & \parbox{0.8cm}{Action} & \parbox{0.7cm}{Verb} & \parbox{0.7cm}{Noun} & \parbox{0.8cm}{Action}\\
		\hline 
		\multirow{5}{*}{\rotatebox[origin=t]{90}{S1}} & 2SCNN (FUSION)~\cite{twoStream} & 42.16 & 29.14 & 13.23 & 80.58 & 53.70 & 30.36 & 29.39 & 30.73 & 5.35 & 14.83 & 21.10 & 4.46\\ \cline{2-14}
		& TSN (RGB)~\cite{tsn} & 45.68 & 36.80 & 19.86 & 85.56 & 64.19 & 41.89 & \textbf{61.64} & 34.32 & 9.96 & 23.81 & 31.62 & 8.81\\ \cline{2-14}
		& TSN (FLOW)~\cite{tsn} & 42.75 & 17.40 & 9.02 & 79.52 & 39.43 & 21.92 & 21.42 & 13.75 & 2.33 & 15.58 & 9.51 & 2.06\\ \cline{2-14}
		& TSN (FUSION)~\cite{tsn} & 48.23 & 36.71 & 20.54 & 84.09 & 62.32 & 39.79 & 47.26 & 35.42 & 10.46 & 22.33 & 30.53 & 8.83\\ \cline{2-14}
		& HF-TSN (RGB) & \textbf{57.57} & \textbf{39.9} & \textbf{28.09} & \textbf{87.83} & \textbf{65.37} & \textbf{48.63} & 49.12 & \textbf{35.83} & \textbf{11.38} & \textbf{39.37} & \textbf{37.04} & \textbf{13.84}\\
		\hline \hline
		\multirow{5}{*}{\rotatebox[origin=t]{90}{S2}} & 2SCNN (FUSION)~\cite{twoStream} & 36.16 & 18.03 & 7.31 & 71.97 & 38.41 & 19.49 & 18.11 & 15.31 & 2.86 & 10.52 & 12.55 & 2.69\\ \cline{2-14}
		& TSN (RGB)~\cite{tsn} & 34.89 & 21.82 & 10.11 & 74.56 & 45.34 & 25.33 & 19.48 & 14.67 & 4.77 & 11.22 & 17.24 & 5.67\\ \cline{2-14}
		& TSN (FLOW)~\cite{tsn} & 40.08 & 14.51 & 6.73 & 73.40 & 33.77 & 18.64 & 19.98 & 9.48 & 2.08 & 13.81 & 8.58 & 2.27 \\ \cline{2-14}
		& TSN (FUSION)~\cite{tsn}& 39.4 & 22.70 & 10.89 & 74.29 & 45.72 & 25.26 & 22.54 & 15.33 & 5.60 & 13.06 & 17.52 & 5.81\\ \cline{2-14}
		& HF-TSN (RGB) & \textbf{42.40} & \textbf{25.23} & \textbf{16.93} & \textbf{75.76} & \textbf{48.96} & \textbf{33.32} & \textbf{24.25} & \textbf{20.48} & \textbf{6.29} & \textbf{15.77} & \textbf{21.96} & \textbf{10.05}\\ \hline
	\end{tabular}
	\caption{Comparison of recognition accuracies with state-of-the-art on EPIC-KITCHENS dataset. HF-TSN is trained for verb, noun and action classification.}
	\label{tab:epic_kitchens}

\end{table}

{\noindent\textbf{EPIC-KITCHENS:}} HF-TSN model is compared against the baseline methods on Tab.~\ref{tab:epic_kitchens}. Results on all other models are provided by the dataset developers~\cite{damen2018scaling}. The model is trained for verb, noun and action classification. We also apply the action prediction score as a bias to the verb and noun classifiers. We report the scores on the test set obtained from the evaluation server. It can be seen that HF-TSN obtained a gain of $12\%$ and $8\%$ for verb classification on S1 and S2 settings over \ac{tsn} (RGB) baseline. In fact, HF-TSN from RGB surpasses the performance of the baselines that use both RGB and optical flow, showing its capacity for extracting highly discriminative short and long term spatio-temporal features. 

\section{Conclusions}
\label{sec:conclusions}
We presented a hierarchical aggregation scheme for video understanding architectures with CNN backbone that is lightweight and effective. In HF-Nets, adjacent feature branches interact between feature differencing and averaging as they compile higher level representations, thereby providing cheap spatio-temporal features for competitive performance.
We plugged HF on top of different baseline models (TSN, TRN, ECO) and evaluated on three public action recognition datasets, obtaining consistent performance improvements. We improve action recognition accuracy of TSN on video clips of complex human-object relationships~\cite{goyal2017something} by more than 24\% while introducing only about 1\% additional trainable parameters and 1\% FLOPs of computation overhead in inference. 
Since HF-Net scheme can be plugged into any deep video architecture with CNN backbone, our future work includes the evaluation of additional baselines and two-stream solutions. 


\bibliographystyle{ieeetr}
\bibliography{bmvc_refs}
\end{document}